\documentclass[runningheads]{llncs}
\usepackage{amsmath}
\usepackage{amssymb}
\usepackage{amsmath}
\usepackage{multirow}

\usepackage{graphicx}

\usepackage{hyperref}

\begin{document}
\title{Variational Bayesian sequence-to-sequence Networks for Memory-Efficient Sign Language Translation}

\author{Harris Partaourides\inst{1} \and
Andreas Voskou\inst{1} \and
Dimitrios Kosmopoulos\inst{2} \and \\
Sotirios Chatzis\inst{1} \and
Dimitris N. Metaxas\inst{3}}

\authorrunning{H. Partaourides et al.}
\titlerunning{Variational Bayesian Networks for Memory-Efficient SLT}

\institute{Cyprus University of Technology, Limassol, Cyprus \\
\email{\{c.partaourides, sotirios.chatzis\}@cut.ac.cy, ai.voskou@edu.cut.ac.cy} \\
\and
University of Patras, Greece 
\email{dkosmo@upatras.gr}
\and
Rutgers University, New Jersey, USA 
\email{dnm@cs.rutgers.edu}}

\maketitle              

\begin{abstract}

Memory-efficient continuous Sign Language Translation is a significant challenge for the development of assisted technologies with real-time applicability for the deaf.
In this work, we introduce a paradigm of designing recurrent deep networks whereby the output of the recurrent layer is derived from appropriate arguments from nonparametric statistics. 
A novel variational Bayesian sequence-to-sequence network architecture is proposed that consists of a) a full Gaussian posterior distribution for data-driven memory compression and b) a nonparametric Indian Buffet Process prior for regularization applied on the Gated Recurrent Unit non-gate weights.
We dub our approach Stick-Breaking Recurrent network and show that it can achieve a substantial weight compression without diminishing modeling performance.

\end{abstract}

\section{Introduction}

National Sign Languages (SLs) are an important part of the European and the world cultural diversity.
In Europe, there are 30 official SLs, 750 thousand deaf sign language users but only 12 thousand interpreters. 
This shortage undermines the right for equal education, employment, information, health services \cite{dreuw2010signspeak} and in some cases endangers the lives of deaf people \cite{murray2013lack}.
The scientific community has long been interested in developing assistive technologies for deaf users, but their broad applicability is yet to be achieved. 

Sign Language Recognition (SLR) applications have achieved satisfactory performance on recognizing distinct signs by means of computer vision models.
In contrast, a more demanding and useful in real-life scenarios task is the translation of sign language sentences i.e. continuous Sign Language Translation (SLT).
Efficient continuous SLT requires the seamless combination of computer vision and Neural Machine Translation (NMT) architectures. 
Moreover, the task has an inherent difficulty due to the fact that we are translating between two different mediums of communication.
In other words, SLs are visual languages, while written languages are auditory.
Currently, the state-of-the-art approaches for SLT utilize the encoder-decoder sequence-to-sequence architectures \cite{nlst}.
The encoder extracts the salient information from a signer's video by means of deep convolutional neural networks, and the decoder utilizes the extracted sequential information to produce full-fledged sentences.
Researchers consider SLT a two-fold process, where the first fold considers SLR as an intermediate tokenization component that extracts glosses (signs notations) from video namely Sign2Gloss.
The second part utilizes glosses to perform NMT; the Gloss to Text task namely Gloss2Text.
In this paper, we focus on the latter.

Currently, SLT is considered an open problem with many challenges, especially on the domain of real-time applicability.
The current state-of-the-art approaches entail millions of parameters and suffer from a common issue in Deep Neural Networks (DNNs), namely overparameterization.
This is observed when a large fraction of the network parameters is redundant for modeling performance, and inflicts an unnecessary computational burden that hinders real-time operation on consumer-grade hardware devices, such as mobile phones and laptops.
Moreover, parameter redundancy renders DNNs susceptible to overfitting hence sub-optimal generalization. 

Numerous approaches have been developed by the deep learning community to address overfitting in DNNs, such as $\ell_2$ regularization, Dropout, DropConnect and variational variants thereof \cite{Gal2015DropoutB,partaourides2017deep}. 
To effectively train the network weights, these approaches tackle overfitting through the scope of regularization.
However, since they retain the network weights in their entirety, they do not directly address memory efficiency. 
One popular approach that achieves memory efficient networks is teacher-student networks, where a condensed network (student) is trained by leveraging a larger network (teacher) ~\cite{dodeep,Distilling}.
However, this approach has two main drawbacks a) the joint computational costs of training two networks and b) the increased number of heuristics for effective teacher distillation. 
Alternatively, several authors have proposed data-driven weight compression and regularization on a single trained network.

In this context, Bayesian Neural Networks (BNNs) provide a solid inferential framework to achieve effective regularization of DNNs~\cite{Graves2011}.
A fully probabilistic paradigm is obtained by imposing a prior distribution over the network components, inferring the appropriate posteriors, and obtaining predictive distributions via marginalization in a Bayesian sense.
Since Bayesian inference requires drawing samples from the posterior distribution, we can additionally utilize the posterior variance magnitude to adjust floating-point precision.
This stems from the insight that higher variance implies that lower floating-point precision is required to retain accuracy at the time of Bayesian averaging ~\cite{Welling,panousis2019nonparametric}.
Additionally, as it has been shown in~\cite{Chatzis18}, the use of an additional set of auxiliary Bernoulli latent variables can obtain a sparsity-inducing behavior. 
This can be achieved by imposing appropriate stick-breaking priors~\cite{Ishwaran2001} over the postulated auxiliary latent variables to explicitly indicate the utility of each component and perform data-driven regularization.

Drawing upon these insights, we propose a principled approach to reduce the network memory footprint by employing nonparametric statistics on the recurrent component of the sequence-to-sequence networks employed for SLT.
We achieve this by imposing a full Gaussian posterior distribution on the Gated Recurrent Unit (GRU) non-gate weights that offers a natural way for weight compression.
Additionally, we impose a nonparametric Bayesian Indian Buffet Process (IBP) prior \cite{Griffiths05} again on the GRU non-gate weights for regularization.
For efficient training and inference of the network, we rely on stochastic gradient variational Bayes.
We dub our approach Stick-Breaking Recurrent (SB-GRU) network, and evaluate its effectiveness on the Gloss-to-Text SLT task, making use of the PHOENIX-2014T benchmark dataset \cite{nlst}. 

We provide a number of quantitative results that exhibit the ability of our approach to yield improved predictive accuracy, in parallel with appropriate data-driven schemes to perform weight compression and regularization. Furthermore, our approach produces faster inference thus taking SLT a step closer to real-time applicability on memory constrictive devices.

The rest of this paper is organized as follows: 
In Section 2, we introduce the necessary theoretical background. 
In Section 3, we introduce the proposed approach and describe the training and inference algorithms of the model. 
In Section 4, we perform experimental evaluations of our approach and provide insights into its functionality. 
Finally, in the concluding Section, we summarize the contribution of this work, and discuss directions for further research.

\section{Related Work}

In this paper, we investigate memory-efficient translation of sign language to text.
We therefore first review recent advancements in the field of NMT. Thereafter, we briefly present a core 
methodological component used in this work, namely the (nonparametric) Indian Buffet Process process.

\subsection{Neural Machine Translation}

NMT utilizes sequence-to-sequence architectures built upon Recurrent Neural Network (RNN) variants to learn the statistical model for efficient translations. 
In the seminal papers of \cite{cho2014properties,sutskever2014sequence} the authors showed the effectiveness of these types of model architectures in translating different languages.
Initially, simple RNNs were used to learn the translation of a source sequence to a target sequence.
They were soon followed by variations of RNNs that tackle the limited temporal attention span of RNNs e.g. Long Short-Term Memory \cite{hochreiter1997long} and GRU \cite{Chung2014}. 
These approaches have managed to improve translation performance and achieve longer temporal horizons.
A complementary approach that uses attention mechanisms \cite{bahdanau2015neural} between the encoder and the decoder allows the network to better focus on certain sequences of the input.
This is achieved by a similarity measure between the state of the decoder and the input.
The most used variation of the attention mechanism is the one proposed by Luong et al. \cite{luong2017neural}.
Finally, recent works replace the whole RNN architecture with purely attention-based architectures \cite{vaswani2017attention}.

Typically, sequence-to-sequence models include an initial embedding layer for vector representation of input words.
This representation of words lies in a high-dimensional plane where similar words are placed closer together.
This approach is preferred than the simplistic one-hot encoding procedure.
However, in the context of SLT, and in order to accommodate visual features, this embedding is replaced with a feature extraction process from video using Convolutional Neural Networks (CNNs) \cite{camgoz2017subunets}.
To better capture spatio-temporal relationships, 3D CNNs \cite{guo2018hierarchical} can be further utilized.
Additionally, several research works have taken advantage of glosses information pertinent to SLT to facilitate model training.
This is placed after the computer vision component (e.g. CNN component) as an auxiliary source of information.

\subsection{The Indian Buffet Process}

The Indian Buffet Process (IBP) \cite{Griffiths05} is a probability distribution over infinite sparse binary matrices. 
It is a stochastic process suitable to be used as a flexible prior, allowing the number of considered features to increase as more data points are made available.
Using the IBP, we can infer in a data-driven fashion a finite number of features for modeling a finite set of observations, in a way that ensures sparse representations \cite{theo}. 
These features can be expanded as new observations appear.
To be used in Variational Inference the authors in \cite{TehGorGha2007} presented a stick-breaking construction for the IBP.

Let us consider $N$ observations, $\boldsymbol{Z}=[z_{j,k}]_{j,k=1}^{N,K}$ denotes a binary matrix where $j$ an observation and $k$ a feature. 
The binary matrix indicates the existence of the features in the observations.
As features approach infinity, $K\rightarrow \infty$, we arrive at the following hierarchical representation for the IBP \cite{TehGorGha2007}:
\begin{equation}
\begin{aligned}[t]
u_k &\sim \text{Beta}(\alpha,1)
\end{aligned}
\quad 
\begin{aligned}[t]
\pi_k = \prod_{j=1}^k u_j \\
\end{aligned}
\quad
\begin{aligned}[t]
z_{jk}&\sim \text{Bernoulli}(\pi_k)\quad  
\end{aligned}
\end{equation}
where $u$ the stick variables and $\alpha>0$ is the innovation (or strength) hyperparameter of the process, a non-negative parameter that controls the magnitude of the induced sparsity. 
In practice, we limit $K$ to the input dimensionality to avoid overcomplete representation of the observed data.

\section{Proposed Approach}

Let us assume an input dataset $\{\boldsymbol{x}_n\}_{n=1}^N\in\mathbb{R}^{J}$, where $N$ the observations and $J$ the features of each observation.
Deep networks perform input representation to extract informative features from the raw input data by means of hidden layers with nonlinear units. 
To produce layer outputs $\{\boldsymbol{y}_n\}_{n=1}^N\in\mathbb{R}^{K}$, the model performs affine transformations via the inner product of the input with the layer weights $\boldsymbol{W}\in\mathbb{R}^{J\times K}$, where K the number of output units of the layer. 
In each example $n$, the derived input representation yields $\boldsymbol{y}_n = \sigma (x_n\boldsymbol{W} + b)$ where $b\in\mathbb{R}^{K}$ the bias and $\sigma (\cdot)$ a non-linear activation function.
These outputs are used as inputs to the next layer.

For data-driven weight compression and regularization, we adopt concepts from Bayesian nonparametrics.
For regularization, we retain the needed synaptic weights by employing a matrix of binary latent variables $\boldsymbol{Z} \in \{0,1\}^{J\times K}$, where each entry therein is $z_{j,k}=1$, if the synaptic connection between the $j^{th}$ dimension of the input and the $k^{th}$ feature is retained, and $z_{j,k}=0$ otherwise.
To perform inference, we impose the sparsity-inducing IBP prior over the binary latent variables $\boldsymbol{Z}$.
The corresponding posterior distribution is independently drawn from $q(z_{j,k}) = \text{Bernoulli}(z_{j,k} | \tilde{\pi}_{j,k})$.
This promotes retention of the barely needed connections based on the theory detailed in Sec. 2.2.
For weight compression, we define a distribution over the synaptic weight matrices, $\boldsymbol{W}$. 
A spherical prior $\boldsymbol{W} \sim \prod_{j,k} \mathcal{N}(w_{j,k}| 0, 1)$ is imposed for simplicity purposes and the corresponding posterior distribution to be inferred $q(\boldsymbol{W}) = \prod_{j,k} \mathcal{N}(w_{j,k} | \mu_{j,k},\sigma_{j,k}^2)$. 
The resulting dense layer output takes the following form:
\begin{align}
[\boldsymbol{y}_n]_{k} &= \sigma ( \sum_{j=1}^J (w_{j,k} \cdot z_{j,k} ) \cdot [\boldsymbol{x}_{n}]_j + b_j ) \; \in \mathbb{R} 
\end{align}

This concludes the formulation of a dense layer with IBP prior over the utility indicators and spherical prior over the synaptic weights. We employ this mechanism to effect data-driven weight compression and regularization, as we show below.



\subsection{A Recurrent Variant}

To accommodate architectures that utilize recurrent connections, we adopt the following rationale. 
Let us assume an input tensor $\{\boldsymbol{X}_n\}_{n=1}^N\in\mathbb{R}^{T\times J}$ where $N$ the observations, $T$ the timesteps and $J$ the features of each observation.
At each timestep, the recurrent layer output  $\{\boldsymbol{y}_n\}_{n=1}^N\in\mathbb{R}^{K}$ is derived by means of the input $\boldsymbol{x}_{t,n}$ and the previous output $\boldsymbol{y}_{t-1}$, where $K$ is the number of output units of the layer. 
Hence, for each example $n$, at each timestep $t$, the input representation yields:
$\boldsymbol{y}_{t,n} = \sigma (x_{t,n}\boldsymbol{W}_{input} + y_{t-1}\boldsymbol{W}_{rec} + b)$ with weights $\boldsymbol{W}_{input}\in\mathbb{R}^{J\times K}$, $\boldsymbol{W}_{rec}\in\mathbb{R}^{K\times K}$, bias $b\in\mathbb{R}^{K}$ and $\sigma (\cdot)$ a non-linear activation function.
We can, then, employ the utility latent indicator variables, $z$, and a distribution over the synaptic weight matrices over any weight matrix, via a procedure similar to the aforementioned dense-layer formulation.

In this paper, we consider the Gated-Recurrent Unit (GRU)~\cite{cho2014learning} that reads:
\begin{equation}
m_t = \sigma ([y_{t-1},x_t]\boldsymbol{W}_{J+K,K} + b_m)
\end{equation}
\begin{equation}
r_t = \sigma ([y_{t-1},x_t]\boldsymbol{W}_{J+K,K} + b_r)
\end{equation}
\begin{equation}
\overset{\sim}y_t = \tanh([r_t \odot y_{t-1},x_t]\boldsymbol{W}_{J+K,K} + b_y)
\end{equation}
\begin{equation}
y_t = (1 - m_t) \odot y_{t-1} + m_t \odot \overset{\sim}y_t
\end{equation}
where $[\cdot,\cdot]$ the concatenation between two vectors, $\odot$ the Hadamard product and $\boldsymbol{W}_{J+K,K}$ the concatenation between $\boldsymbol{W}_{input}$ and $\boldsymbol{W}_{rec}$.
We apply our proposed modeling principles and rationale only on the non-gate related weights. We dub our recurrent layer variant the Stick-Breaking Gated Recurrent Unit (SB-GRU).
Eq.5 now reads:
\begin{equation}
\overset{\sim}y_t = \tanh([r_t \odot y_{t-1},x_t]\boldsymbol{W}_{J+K,K}\boldsymbol{Z}_{J+K,K} + b_y)
\end{equation}

\subsection{Model Training}

The proposed model can be trained by using Stochastic Gradient Variational Bayes (SGVB) to maximize its Evidence Lower Bound (ELBO). 
As the resulting ELBO expression cannot be computed analytically, we resort to Monte-Carlo sampling techniques that ensure low-variance estimation. These comprise the standard reparameterization trick, the Gumbel-SoftMax relaxation trick \cite{maddison2017concrete}, and the Kumaraswamy reparameterization trick \cite {Kumaraswamy1980}, applied to the postulated Gaussian weights $\boldsymbol{W}$, the discrete latent indicator variables $\boldsymbol{Z}$, and the stick variables $\boldsymbol{u}$, respectively. 
The standard reparameterization trick applied on the Gaussian weights cannot be used for the IBP prior Beta-distributed stick variables.
Thus, one can approximate the variational posteriors $q(u_k)=\mathrm{Beta}(u_k|a_k,b_k)$ via the Kumaraswamy distribution \cite{Kumaraswamy1980} that reads:
\begin{align}
q(u_k;a_k,b_k) &= a_k b_k u_k^{a_k-1}(1-u_k^{a_k})^{b_k-1}
\end{align}
where samples from $q(u_k;a_k,b_k)$ can be reparametrized as follows \cite{nali}:
\begin{align}
u_k = 
\left( 1 - (1-X)^{\frac{1}{b_k}}\right)^{\frac{1}{a_k}}, \; X \sim U(0,1)
\end{align}


We know turn to the case of the Discrete Bernoulli latent variables of our model.
Performing back-propagation through reparametrized drawn samples is infeasible, hence recent solutions introduced appropriate continuous relaxations \cite{Jang2017} \cite{maddison2017concrete}.
Let us assume a Discrete distribution $\boldsymbol{X}=[X_k]_{k=1}^{K}$.
The drawn samples are of the form:
\begin{align}
\label{eqn:concrete_sampling}
X_k &= \frac{\exp(\log \eta_k+G_k)/\lambda)}{\sum_{i=1}^K \exp((\log \eta_i+G_i)/\lambda)}, \\ G_k &= -\log(-\log U_k), \; U_k \sim\text{Uniform}(0,1)
\end{align}
where $\boldsymbol{\eta} \in (0,\infty)^K$ the \emph{unnormalized} probabilities and $\lambda \in (0,\infty)$ the \textit{temperature} of the relaxation of the differentiable functions.
In this paper, we anneal the values of the hyperparameter $\lambda$ similar to \cite{Jang2017}.
We further consider posterior independence a) across layers and, b) among the latent variables $\boldsymbol{Z}$ in each layer. 
All the posterior expectations in the ELBO are computed by drawing MC samples under the Normal, Gumbel-SoftMax and Kumaraswamy reparameterization tricks, respectively.
Hence, the resulting ELBO is of the form:
\begin{align}
\label{eqn:elbo_full}
\begin{split}
\mathcal{L}(\phi) = \mathbb{E}_{q(\cdot)}\Big[\log p(\mathcal{D}|\boldsymbol{ Z,  u, W})  - \Big(& KL \big[\ q(\{ \mathbf{Z} \}) \ || \ p(\{ \mathbf{Z}| \boldsymbol{u} \})\ \big] \\ & + KL \big[\ q(\{\boldsymbol{u}\}) \ || \ p(\{\boldsymbol{u}\})\ \big] \\ & + KL \big[\ q(\{ \boldsymbol{W}\}) \ || \ p(\{\boldsymbol{W}\})\ \big] \Big)\Big]
\end{split}
\end{align}
where the KL divergences of the stick-variables $u_k$, weight utility indicators $z_{j,k}$, and Gaussian weights can be obtained by:
\begin{align}
\label{eqn:kl_u}
\begin{split}
KL[q(u_k)|p(u_k)] &= \mathbb{E}_{q(u_k)}[\log p(u_k) - \log q(u_k)] \\
&\approx \log p(\hat{u_k}) - \log q(\hat{u_k}), \ \forall k
\end{split}
\end{align}
\begin{align}
\begin{split}
KL[q(z_{j,k})|p(z_{j,k})] &= \mathbb{E}_{q(z_{j,k}), q(u_k)}[\log p(z_{j,k}|\pi_k) - \log q(z_{j,k})]\\
&\approx  \log p(\hat{z}_{j,k}|\prod_{i=1}^k \hat{u_i}) - \log q(\hat{z}_{j,k}), \ \forall j,k
\end{split}
\end{align}
\begin{align}
\label{eqn:kl_w}
\begin{split}
KL[(q(w_{j,k,u})|p(w_{j,k,u})] &= \mathbb{E}_{q(w_{j,k,u})}[\log p(w_{j,k,u}) - \log q(w_{j,k,u})]\\
&\approx  \log p(\hat{w}_{j,k,u}) - \log q(\hat{w}_{j,k,u}), \ \forall j,k,u
\end{split}
\end{align}
We can then employ standard off-the-self stochastic gradient techniques to maximize the ELBO.
In our experiments, we adopt ADAM \cite{kingma2014adam}.

\subsection{Inference Algorithm}

Having trained our SBR model, we can now use the posteriors to perform inference for unseen data. 
Our model exhibits two advantages compared to conventional techniques.
First, exploiting the inferred weight utility latent indicator variables, we can omit the contribution of weights that are effectively deemed unnecessary in a data-driven fashion. 
Specifically, we employ a \emph{cut-off threshold}, $\tau$ to omit any weight with inferred corresponding posterior $q(z)$ below $\tau$. 
Typically, in Bayesian Neural Networks literature the utility is only \emph{implicitly inferred} by means of \emph{thresholding higher-order moments}.
These thresholds are derived on the hierarchical densities of the \emph{network weights themselves}, $\boldsymbol{W}$ and require extensive heuristics for the selection of the hyperparameter values in each network layer.
Our approach is in direct contrast to these techniques where we only need to specify the global innovation hyperparameter $\alpha$ and global truncation threshold $\tau$.
Moreover, our model formulation is robust to small fluctuations of their values. 

Second, exploiting the inferred variance of the Gaussian posterior distribution employed over the network weights, $\boldsymbol{W}$, we can reduce the weights floating-point bit precision level in a data-driven fashion.
Specifically, we employ a unit round off to limit the memory requirements for representing the network weights.
This is achieved by means of the mean of the weight variance \cite{Welling}.
In principle, the higher the level of uncertainty present on the Gaussian posterior the less bits are actively contributing to inference since their approximate posterior sampling fluctuates too much.
In contrast to \cite{Welling}, we disentangle the two procedures (bit-precision and omission) by imposing \emph{different posteriors}. 
This alleviates the tendency to underestimate posterior variance, therefore better retaining predictive performance while performing stronger network compression.

In a proper Bayesian setting, at the prediction generation stage we need to perform averaging of multiple samples which is inefficient. 
Without losing generality, a common approximation is the replacement of weight values with their weight posterior means during forward propagation \cite{Welling,Neklyudov}.
To achieve weight compression, we replace the trained weights with their bit-precision counterparts.

\section{Experimental Evaluation}

To exhibit the efficacy and effectiveness of our approach, we perform a thorough experimental evaluation in the context of an SLT task.
We use the model architecture from \cite{nlst}, which consists of 4 layers in the encoder and 4 in the decoder, with 1000 units in each layer; we replace the last layer of the encoder with our proposed recurrent layer variant.
The task we consider for our experiments is the Gloss to Text Translation.
To this end, we utilize the primary benchmark for SLT, namely “RWTH-PHOENIX-Weather 2014T” dataset \cite{nlst}.
Additionally, we also evaluate the Gloss2Text GRU model without attention, from the reference paper, as a baseline.
In all cases, we perform stochastic gradient descent by means of Adam \cite{kingma2014adam}, with learning rate 0.00001, batch size 128 and dropout 0.2, until convergence.
The performance metrics used in our experiments are BLUE and ROUGE. 
We implement our model in TensorFlow \cite{tensorflow2015-whitepaper}. 

\subsection{RWTH-PHOENIX-Weather 2014T Dataset}

PHOENIX14T is a collection of weather forecast videos in German sign language, annotated with the corresponding glosses and fully translated in German text.
This annotation process makes it directly applicable to Sign to Text, Sign to Gloss and Gloss to Text translation.
The glosses annotation procedure has been done manually by deaf experts, and the text has been automatically transcribed via available Speech to Text software, directly applied on the spoken weather forecast.

In this paper, we consider the Gloss to Text subset of the dataset. 
This includes 8257 pairs of gloss to text sentences where 7096 are used for training, 519 for validation and 642 for testing. 
Glosses represent the discrete gestures of the sign language in textual form. 
Additionally, auxiliary words are used to stand for non-manual information that go with the signs. 
This results in a simplified sign language vocabulary compared to the spoken language counterpart, but augmented with auxiliary words.
In the specific corpus, we have a glosses vocabulary size of 1066 (337 singletons) and a German vocabulary of 2887 (1077 singletons).

\subsection{Quantitative Study}

The SB-GRU model architecture comprises of two new processes: weight reparameterization and stick-breaking process.
We perform an ablation study on these components to exhibit the effectiveness of each mechanism on the model performance.
In this context, we run our experiments with the mechanisms separately and in conjunction.
We dub GRU$_{repar}$ and GRU$_{BP}$ the reparameterization and stick-breaking only process model variant and compare with the Gloss2Text GRU model without attention used in \cite{nlst}.
Moreover, we include the performance of the models when applied with weight compression, GRU$_{repar,wc}$ and SB-GRU$_{wc}$
The results are shown in Table 1.

\begin{table}
\centering
\caption{Performance Metrics }\label{table:tab1}
\begin{tabular}{|l|c|c|c|c|c|c|}
\hline
\multicolumn{1}{|l|}{\multirow{2}{*}{Model}} & \multicolumn{2}{c|}{Dev} & \multicolumn{2}{c|}{Test} \\ 
\cline{2-5} \multicolumn{1}{|l|}{} & \multicolumn{1}{l|}{Bleu-4} & \multicolumn{1}{l|}{Rouge} & \multicolumn{1}{l|}{Bleu-4} & \multicolumn{1}{l|}{Rouge} \\ 
\hline
Baseline & 16.3 & 40.3 & 16.3 & 40.7 \\
GRU$_{repar}$& 16.7 & 41.1 & 17.0 & 41.5 \\
GRU$_{repar,wc}$& 16.2 & 40.6 & 16.7 & 40.7 \\
GRU$_{bp}$ & 18.4 & 43.9 & 17.0 & 43.1 \\
SB-GRU & 17.9 & 43.0 & 18.1 & 43.5 \\
SB-GRU$_{wc}$ & 17.7 & 43.0 & 17.8 & 42.8 \\
\hline
\end{tabular}
\end{table}

We observe that utilizing an IBP prior over the utility indicators and a spherical prior over the synaptic weights leads to a solid performance increase. 
Separately applied, the two schemes present similar performance improvements; in both cases, the GRU$_{bp}$ exhibit some overfitting.
However, when used in conjunction in the context of the proposed SB-GRU network, we yield a significant performance increase in terms of both the considered performance metrics.
In addition, by performing weight posterior inference, we can reduce the weight memory footprint by a factor of 16, with a required bit precision of 1, without losing significant performance.
Regarding computational complexity, the baseline model performs a training step in 0.3s, GRU$_{repar}$ in 0.33s (+12\%), GRU$_{bp}$ in 0.44s (+46\%) and SB-GRU in 0.48s (+59\%).

\subsection{Qualitative results}

To provide some qualitative insights on the Gloss2Text task, in Table 2 we present few examples from the test set ground truth (glosses + translation), along with the translation samples generated from our various networks.
To facilitate understanding, we also include the English translation of these examples.

\begin{table}[]
\centering
\caption{Predictive performance on the Test set}\label{Exam-Test}
\resizebox{0.9\columnwidth}{!}{%
\begin{tabular}{|l|l|}
\hline
Gloss            & jetzt wetter wie-aussehen morgen samstag zweite april \\ 
Text             & und nun die wettervorhersage für morgen samstag den zweiten april .    \\
Text(eng)        & and now the weatherforecast for tomorrow saturday the second april     \\

\hline
GRU$_{repar}$    & und nun die wettervorhersage für morgen samstag den zweiten april .    \\
GRU$_{repar,wc}$ & und nun die wettervorhersage für morgen samstag den zweiten april .    \\
GRU$_{bp}$       & und nun die wettervorhersage für morgen samstag den zweiten april .    \\
SB-GRU           & und nun die wettervorhersage für morgen samstag den zweiten april .    \\
SB-GRU$_{wc}$    & und nun die wettervorhersage für morgen samstag den zweiten april .    \\
\hline
\hline
Gloss            & montag anfang wechselhaft mehr kuehl \\
Text             & die neue woche beginnt wechselhaft und kuhler .                        \\
Text(eng)        & the new week starts unpredictable and cooler .                         \\
\hline
GRU$_{repar}$    & am montag wieder wechselhaft und kühler .                              \\
GRU$_{repar,wc}$ & auch am montag wechselhaft und wechselhaft .                           \\
GRU$_{bp}$       & die neue woche beginnt wechselhaft und wieder kühler .                 \\
SB-GRU           & die neue woche beginnt wechselhaft und wieder kuhler .                 \\
SB-GRU$_{wc}$    & am montag wieder wechselhaft und kühler .                              \\
\hline
\hline
Gloss            & sonntag regen teil gewitter \\
Text             & am sonntag ab und an regenschauer teilweise auch gewitter.             \\
Text(eng)        & on sunday from time to time rain showers sometimes also thunderstorms. \\
\hline
GRU$_{repar}$    & am sonntag regnet es ab und an .                                       \\
GRU$_{repar,wc}$ & am sonntag regnet es ab und an .                                       \\
GRU$_{bp}$       & am sonntag regnet es im nordwesten einzelne gewitter .                 \\
SB-GRU           & am sonntag verbreitet regen oder gewitter .                            \\
SB-GRU$_{wc}$    & am sonntag verbreitet regen oder gewitter .     \\
\hline
\end{tabular}
}
\end{table}


We observe that the overall grammar of the translation is coherent with major errors present at the translation of places, numbers, and dates.
This is a direct effect of the infrequent words and singletons that exist in the specific dataset.

\section{Conclusions}

Memory-efficient continuous sign language translation is a significant challenge for the development of assisted technologies with real-time applicability.
In this paper, we introduced a novel variational Bayesian sequence-to-sequence network architecture that consists of a) a full Gaussian posterior distribution on the GRU non-gate weights for data-driven memory compression and b) a nonparametric IBP prior on the same weights for regularization.
We dubbed our approach Stick-Breaking Recurrent network and showed that we can achieve a substantial weight compression without diminishing modeling performance/accuracy. 
These findings motivate us to further examine the efficacy of these principles when applied to the more challenging Sign to Text task, specifically their applicability to the convolutional layers of the network, as well as the complete model structure. We also intend to investigate the usefulness of applying asymmetric densities to the weights \cite{partaourides2017asymmetric}.
Finally, another focus area for future work is the employment of quantum-statistical principles for model training \cite{chatzis2012quantum}, that may facilitate better handling of the uncertainty that stems from the limited availability of training data.

\section*{Acknowledgments}
This research was partially supported by the Research Promotion Foundation of Cyprus, through the grant: INTERNATIONAL/USA/0118/0037.

\bibliographystyle{splncs04}
\bibliography{bibliography}
\end{document}